\PassOptionsToPackage{OT2,TS1,T1}{fontenc}
\documentclass{ceurart}

\usepackage[utf8]{inputenc}
\usepackage{textcomp}

\newcommand{\longs}{{\fontencoding{TS1}\fontfamily{lmr}\selectfont\char115}}

\newcommand{\yat}{{\fontencoding{OT2}\fontfamily{cmr}\selectfont\cyryat}}
\newcommand{\fita}{{\fontencoding{OT2}\fontfamily{cmr}\selectfont\cyrfita}}
\newcommand{\izhitsa}{{\fontencoding{OT2}\fontfamily{cmr}\selectfont\cyrizh}}

\usepackage{listings}
\begin{document}

\copyrightyear{2026}
\copyrightclause{Copyright for this paper by its authors.
  Use permitted under Creative Commons License Attribution 4.0
  International (CC BY 4.0).}

\conference{The 22nd Conference on Information and Research Science Connecting to Digital and Library Science}

\title{How Surprising Is Historical Italian to Language Models? Tokenization Tax, Comprehension Tax, and a Simple Mitigation}

\author[1]{Maria Levchenko}[%
orcid=0009-0008-2649-0729,
email=mariia.levchenko@unibo.it,
url=https://mary-lev.github.io,
]
\cormark[1]
\fnmark[1]
\address[1]{University of Bologna}

\begin{abstract}
  Large language models (LLMs) are increasingly critical to digital library workflows, yet their ability to process historical language remains poorly understood. Historical difficulty is typically treated as a monolithic barrier, conflating orthographic variation, linguistic distance, and pretraining exposure. In this paper, we propose a diagnostic framework that decomposes this difficulty into four distinct dimensions: tokenization cost, predictive uncertainty (surprisal), semantic robustness, and context sensitivity.

  We evaluate this framework on three datasets spanning three centuries: (1) a newly curated corpus of 17th-century Italian texts (1610--1689) digitized from original page images; (2) canonical 19th-century Italian (\textit{I Promessi Sposi}) serving as a high-exposure control; and (3) 18th-century Russian civil print books as a contrastive orthographic stress test.

  Our results reveal a \textbf{distinct} dissociation between encoding cost and comprehension. While Russian and early modern Italian incur comparable tokenization penalties ($\approx$25--30\% inflation), their predictive difficulty diverges sharply. 17th-century Italian is on average 2.4$\times$ more surprising than its modern equivalent---with academic prose reaching 3.2$\times$---whereas Russian shows only a modest increase. But predictive uncertainty does not imply \textbf{representational degradation}: embedding similarity remains robust ($\geq$0.85) across all datasets, confirming that models can represent historical meaning even when generation is unstable.

  Finally, we demonstrate that a minimal temporal context prompt reduces historical surprisal by approximately 60\%, offering a simple, model-agnostic mitigation. These findings suggest that while historical text imposes a consistent encoding tax, digital libraries can safely deploy LLMs for semantic retrieval tasks, provided that generative applications are carefully adapted.
\end{abstract}

\begin{keywords}
historical language \sep
Italian historical texts \sep
language models \sep
perplexity \sep
tokenization \sep
semantic embeddings \sep
digital libraries
\end{keywords}

\maketitle

\section{Introduction}

Large language models (LLMs) are increasingly deployed in digital library workflows, including optical character recognition (OCR) post-correction, indexing, retrieval, summarization, and modernization of historical texts. Yet evaluating how such models handle historical language remains challenging. ``Historical difficulty'' is often treated as a single phenomenon, conflating orthographic variation, linguistic distance, and potential familiarity effects arising from model pretraining.

In this paper, we approach historical language through the lens of \emph{surprisal}: how unexpected a text is to a modern language model, measured via perplexity. This framing allows us to distinguish between \emph{encoding costs} (tokenization inefficiency) and \emph{comprehension costs} (predictive uncertainty), which are often implicitly assumed to coincide. We refer to these as the ``tokenization tax'' and ``comprehension tax,'' respectively — systematic, hidden costs that historical text properties impose on different stages of model processing.

A central methodological challenge is \emph{exposure}. Canonical historical works—widely digitized, edited, and circulated—are likely over-represented in the training distributions of modern LLMs. Low perplexity on such texts should therefore not be interpreted as evidence of robust historical competence. Instead, such works can serve as \emph{best-case historical baselines}, illustrating how models behave when orthography is standardized and lexical material is plausibly familiar.

We adopt this perspective in our treatment of Alessandro Manzoni’s \textit{I Promessi Sposi}. While the novel is historically marked, it occupies a central position in the Italian literary canon and is extensively available in digital form. In the main analysis, we therefore use only Manzoni’s standardized \textit{Quarantana} (1840–42) edition, which exhibits near-modern behavior: negligible tokenization overhead and only marginal increases in perplexity relative to contemporary Italian. We treat Manzoni not as representative of historical Italian in general, but as a high-exposure historical control condition, providing an upper bound on expected model performance for historical text.

To probe genuinely challenging historical language, we contrast this baseline with two datasets that introduce increasing distance from modern usage. 
First, we analyze a newly curated corpus of five 17th-century Italian texts (1610--1689), spanning genres from religious treatises to academic prose. These texts are characterized by the long s (\textit{\longs}), archaic vocabulary, and Latin-influenced syntax.
Second, we include 18th-century Russian civil print books as a \emph{contrastive control condition}: Russian exhibits extreme orthographic divergence and severe tokenization disruption, but comparatively limited lexical and syntactic distance from its modern form. This contrast allows us to disentangle orthographic encoding effects from deeper comprehension difficulty.

While we cannot audit proprietary pretraining corpora, the absence of readily available full-text transcriptions for these specific 17th-century editions reduces the likelihood of direct verbatim exposure and allows us to study historical surprisal under more demanding conditions.

Across these datasets, we ask three questions:
(1) How much does historical text increase tokenization cost?
(2) How surprising is historical text to modern language models?
(3) Can simple, model-agnostic interventions mitigate this difficulty?

This study makes three primary contributions to the evaluation of LLMs in digital libraries:
\begin{enumerate}
    \item We demonstrate that tokenization efficiency is a mechanical tax uncorrelated with model understanding. While 18th-century Russian and 17th-century Italian share similar encoding costs ($\approx$30\% inflation), their predictive difficulty diverges by over 2$\times$, indicating that subword fertility is a poor proxy for historical complexity.
    \item We identify a systematic dissociation where models struggle to generate historical text (high perplexity) while successfully preserving its semantics (high embedding similarity). This validates the safety of using modern LLMs for historical \emph{retrieval} tasks even when \emph{generation} is unstable.
    \item We show that a minimal temporal context prompt reduces historical surprisal by approximately 60\%, offering a cost-effective method to improve model alignment without fine-tuning.
\end{enumerate}

\section{Background \& Related Work}

Historical difficulty for Large Language Models (LLMs) is not a single phenomenon but an interaction of distinct pressures: \emph{encoding costs} introduced by tokenization, \emph{distributional shift} that increases predictive uncertainty, and \emph{temporal misalignment} that biases model expectations. Our study builds on methodological frameworks for evaluating LLMs in digital library and OCR settings, where historical language acts as a recurring stressor \cite{levchenko2025histcorpora,levchenko2025ocrframework,smith2007tesseract}.

\textbf{Tokenization and the ``Orthographic Tax.''} 
Most modern LLMs rely on data-driven subword tokenization (BPE, Unigram) optimized for contemporary corpora \cite{sennrich2016bpe,kudo2018sentencepiece}. When applied to historical text, these tokenizers exhibit high \emph{fertility}—splitting words into excessive sub-units—which inflates computational cost and reduces context capacity \cite{wegmann2025tokenization,MaksymenkoTuruta2025TokenizationUkrainian}.
Prior work in historical OCR suggests that this fragmentation is often mechanical: it arises from orthographic mismatch (e.g., missing characters or rare sequences) rather than semantic incomprehensibility. However, it remains an open question whether this ``tokenization tax'' serves as a reliable proxy for downstream model difficulty. 

\textbf{Perplexity, Surprisal, and the Exposure Confound.} 
Perplexity operationalizes \emph{surprisal} as next-token predictive uncertainty, a measure widely used to estimate the distributional distance between a model's training data and the target text \cite{hale2001surprisal,levy2008expectation}. In diachronic settings, high perplexity typically signals drift in lexicon, syntax, and register.
However, this signal is easily confounded exposure. Large-scale pretraining corpora likely contain canonical historical texts (e.g., literary classics), leading to memorization that masks genuine linguistic difficulty \cite{carlini2021extracting,brown2020gpt3}.
Recent meta-evaluations propose analyzing perplexity signatures to detect such contamination \cite{wu2025mapping}. This motivates our distinction between high-exposure baselines (like Manzoni) and low-resource archival texts for robust evaluation.

\textbf{Semantic Stability vs. Diachronic Shift.} 
While perplexity measures predictive surface alignment, embedding models are typically used to track semantic shift—how word meanings change over time \cite{hamilton2016diachronic,kutuzov2018diachronic}.
In the context of digital libraries, however, the goal is often the inverse: verifying semantic stability. We investigate whether modern embedding models can map archaic surface forms (e.g., \textit{\longs} vs \textit{s}) to the same vector space as their modern equivalents. This approach treats embeddings not as tools for change detection, but as a robustness metric for information retrieval.

\textbf{Temporal Alignment and Prompting.} 
Finally, LLMs frequently lack stable temporal grounding, leading to anachronistic outputs or ``over-historicization'' \cite{levchenko2025histcorpora}.
Recent work suggests that explicit time conditioning (e.g., ``Text from 1687'') can effectively shift model distributions without parameter updates \cite{zhao2024setclocktemporalalignment}.
Consistent with this, we evaluate minimal temporal prompting as a model-agnostic mitigation strategy, testing whether explicitly signaling the temporal domain can bridge the gap between high-cost encoding and semantic understanding.

\section{Datasets}
We used three complementary datasets spanning three centuries of Italian and Russian text. Each dataset is selected to isolate a distinct source of difficulty in historical text processing: orthographic distance, linguistic drift, and plausible exposure.

\paragraph{Russian Civil Print Books (18th Century).} 
The Russian dataset derives from the \textit{Russian Civil Print Books of the 18th Century} collection, a benchmark for historical OCR evaluation {\cite{levchenko2025ocrframework}}. It consists of 374 aligned sentence pairs (historical/modern) from the National Library of Russia, spanning 1752--1801 and covering diverse genres (fiction, religion, history, science).
The texts preserve pre-reform orthography, including obsolete characters such as
\textit{\yat} (yat), \textit{\fita} (fita), and \textit{\izhitsa} (izhitsa),
which are often absent from modern tokenizer vocabularies. This dataset serves as a high-orthographic-distance condition, testing model resilience to byte-level fallback.

\paragraph{Manzoni’s \textit{I Promessi Sposi} (19th Century): High-Exposure Control.} 
For 19th-century Italian, we use the \textit{Quarantana} edition (1840--1842) of Alessandro Manzoni’s \textit{I Promessi Sposi}, digitized by the LeggoManzoni project\footnote{https://projects.dharc.unibo.it/leggomanzoni/} \cite{levchenko2025tei}.
As a central text in the Italian canon, the \textit{Quarantana} is highly likely to be present in LLM pretraining data. We therefore treat it as a \emph{high-exposure historical baseline}.
The dataset includes 400 sentences paired with LLM-generated modernizations, representing a ``best-case'' scenario for historical Italian processing.

\paragraph{17th-Century Italian Corpus (1610--1689).} 
To examine challenging historical Italian, we curated a novel dataset of five texts spanning 79 years, sourced from the FICLIT Digital Library\footnote{https://dlrc.ficlit.unibo.it} (University of Bologna).
Unlike the canonical Manzoni text, these specific editions are served via IIIF manifests and exist primarily as page images, minimizing the likelihood of verbatim pretraining exposure.
The corpus was constructed by performing custom OCR on raw page images. To create the modern baseline, we generated semantic normalizations via \textit{GPT-5-mini}, followed by manual verification to ensure semantic fidelity.
Using a synthetic modern baseline ensures that the control condition reflects a standardized, predictable probability distribution, allowing us to isolate the surprisal caused specifically by historical variation.
Sentences were extracted from continuous page ranges to preserve discourse structure. 
The variation in sample size (e.g., $n$=472 vs. $n$=36) reflects the differing typographical density of the source editions, with the theological treatise (\textit{Della introduttione al simbolo della fede}) exhibiting significantly higher text density per page than the biographical portraits.
In total, the dataset comprises 732 aligned sentence pairs ($\approx$15,000 tokens), preserving dense historical features such as the long s (\textit{\longs}) and V/U interchange.

\begin{table}[t]
\centering
\caption{\textbf{The 17th-Century Italian Corpus.} We selected five texts spanning the century to test the interaction between time and genre. Item IDs refer to the FICLIT Digital Library catalog.}
\label{tab:corpus_metadata}
\small
\begin{tabular}{l l l l r}
\hline
\textbf{Year} & \textbf{Item ID} & \textbf{Short Title} & \textbf{Genre} & \textbf{Pairs} \\
\hline
1610 & 3987 & \textit{Della introduttione al simbolo della fede} & Religious Treatise & 472 \\
1673 & 32066 & \textit{Rime e prose di Claudio Achillini} & Biography & 95 \\
1674 & 32560 & \textit{Dell'arte historica} & Historical Narrative & 54 \\
1687 & 32064 & \textit{Miscellanea poetica} & Academic Preface & 75 \\
1689 & 34771 & \textit{Ritratti historici} & Descriptive Portraits & 36 \\
\hline
\multicolumn{4}{r}{\textit{Total Aligned Sentences}} & \textbf{732} \\
\hline
\end{tabular}
\end{table}

\section{Experimental Framework}

Evaluating historical text requires distinguishing between distinct properties: a text may be expensive to encode, hard to predict, or semantically distant from its modern equivalent. We therefore adopt a multi-metric framework to capture these complementary dimensions of model behavior.

\textbf{Encoding Cost (Tokenization Efficiency).} We quantify the computational burden of historical text by measuring \emph{token inflation}---the relative increase in token counts compared to a modernized baseline. This metric reflects the efficiency of the tokenizer's vocabulary and merge rules, directly determining API costs and effective context window size.

\textbf{Predictive Difficulty (Perplexity).} We measure \emph{surprisal} via perplexity ($PPL$), defined as the exponentiated average negative log-likelihood of a sequence $X = (x_1, \dots, x_T)$:
\begin{equation}
PPL(X) = \exp\left( -\frac{1}{T} \sum_{t=1}^{T} \log P_\theta(x_t \mid x_{<t}) \right)
\end{equation}
where $x_{<t}$ represents the context (preceding tokens).
Since historical texts often fragment into more tokens ($T_{hist} > T_{mod}$), comparing raw likelihoods would be misleading. Perplexity normalizes for sequence length ($T$), making it a per-token measure of uncertainty.
To control for the tokenizer's vocabulary distribution, we report the \emph{perplexity ratio} ($PPL_{hist} / PPL_{mod}$), which isolates the relative difficulty of the historical form compared to its semantic equivalent.
 Unlike tokenization, perplexity captures the model's uncertainty regarding vocabulary, syntax, and register. We interpret high surprisal as a signal of \emph{comprehension difficulty} and distributional mismatch.

\textbf{Semantic Robustness (Embedding Similarity).} To assess whether predictive difficulty implies representational failure, we compute cosine similarity between embeddings of aligned historical and modern sentences. High similarity indicates that the model projects surface variations (e.g., archaic spelling) into a stable semantic space, preserving utility for retrieval and indexing tasks.

\textbf{Contextual Adaptability.} Finally, we isolate the effect of domain mismatch by measuring the reduction in perplexity when a minimal \emph{temporal context prompt} (e.g., ``Italian text from 1687'') is prepended. This metric quantifies how much difficulty is intrinsic to the linguistic form versus how much stems from the model's default lack of temporal grounding.

Together, these four metrics decompose historical difficulty into \emph{cost}, \emph{uncertainty}, \emph{meaning}, and \emph{adaptability}, allowing us to pinpoint exactly where LLMs struggle—and where they remain robust.

\section{Tokenization Analysis}

Tokenization is the first stage of LLM processing and determines how raw text is mapped to discrete input units. Inefficient tokenization increases computational cost, reduces effective context length, and may fragment linguistically meaningful units. Historical texts are often assumed to be difficult for language models precisely because they tokenize poorly. We therefore begin by quantifying the \emph{tokenization cost} of historical text and identifying which historical features contribute most to token inflation.

Tokenization efficiency reflects properties of the tokenizer and character inventory, not necessarily the model’s ability to interpret meaning. A central question for this study is whether tokenization cost can be used as a proxy for broader model difficulty.

We measure tokenization efficiency by comparing token counts for historical text and its modernized equivalent under the same tokenizer. Token inflation is defined as the relative increase in tokens required for the historical version. Our primary analysis uses the OpenAI \texttt{o200k\_base} tokenizer (vocabulary size $\approx$ 200k tokens), which is the standard encoding for the GPT-4o and GPT-5 model families. Where informative, we contrast results with the Qwen~2.5 tokenizer (vocabulary size $\approx$ 152k tokens), which employs a distinct byte-level BPE tailored for stronger multilingual coverage, allowing us to highlight tokenizer-dependent effects.

To isolate the contribution of individual historical features, we apply a cumulative normalization procedure to the 17th-century Italian corpus, progressively correcting specific orthographic and lexical phenomena (e.g., long s, V/U interchange) and measuring token counts after each step.

Across all five 17th-century texts, token inflation remained consistent (Mean: +27.5\%; Range: 26.4\%--30.3\%), despite significant differences in genre and publication date (1610--1689).
This suggests that encoding cost is driven by stable orthographic conventions---primarily the long s (\textit{\longs})---rather than genre-specific register or authorship.

Table~\ref{tab:token_inflation} summarizes tokenization results. It highlights that comparable inflation magnitudes (28--36\%) can arise from distinct mechanisms, distinguishing between texts that are merely expensive to encode and those that are genuinely difficult to predict.

\begin{table}[t]
\centering
\caption{Cross-dataset token inflation under modern tokenizers. Pre-1800 texts incur substantial encoding costs, while canonical 19th-century Italian shows negligible inflation. Note that Italian inflation is driven by subword fragmentation (long s), whereas Russian inflation is driven by byte-level fallback.}
\label{tab:token_inflation}
\begin{tabular}{l l l r r r}
\hline
Dataset & Period & Tokenizer & Historical & Modern & Inflation \\
\hline
Russian civil print & 18th c. & OpenAI & 8,234 & 6,041 & \textbf{+36.3\%} \\
Russian civil print & 18th c. & Qwen & 7,892 & 6,311 & +25.1\% \\
Italian 17th c. Corpus & 1610--89 & OpenAI & 62,870 & 49,314 & \textbf{+27.5\%} \\
\textit{(subset: Miscellanea)} & \textit{1687} & \textit{OpenAI} & \textit{5,288} & \textit{4,125} & \textit{+28.2\%} \\
Manzoni \textit{Quarantana} & 19th c. & OpenAI & 6,509 & 6,510 & $\approx$0\% \\
\hline
\end{tabular}
\end{table}

\paragraph{Failure Modes: Byte Fallback vs. Poor BPE Merges.} Historical token inflation arises through two distinct mechanisms.
In Russian, obsolete characters such as \textit{\yat} and \textit{\fita} are absent from the OpenAI vocabulary and are encoded via byte-level fallback, fragmenting words into multiple tokens. This mechanism produces severe inflation (+36\%) but is largely orthographic in nature.

In 17th-century Italian, by contrast, the dominant factor is the long s (\textit{\longs}). Unlike Russian characters, \textit{\longs} is typically present in tokenizer vocabularies but lacks efficient subword merges due to its rarity in modern training data. As a result, words containing \textit{\longs} are segmented into multiple low-frequency tokens, producing consistent inflation despite full character coverage.

Applying cumulative normalization to the Italian corpus confirms that the long s is the primary driver of this cost. 
Correcting \textit{\longs}$\rightarrow$\textit{s} alone eliminates approximately \textbf{79\%} of the total token inflation (ranging from 70\% to 82\% across individual books). 
The remaining inflation arises primarily from archaic vocabulary and syntactic forms that tokenize less efficiently than their modern equivalents. Other orthographic features (V/U interchange, accents, macrons) contribute negligibly ($<$1\%).
This asymmetry confirms that the ``tokenization tax'' for early modern Italian is largely a mechanical artifact of the long s.

In the following section, we show that this distinction is crucial: texts with comparable tokenization inflation exhibit radically different levels of surprisal, demonstrating that \emph{tokenization cost and comprehension difficulty are not equivalent}.

\section{Perplexity Analysis}

While tokenization efficiency determines the cost of encoding historical text, it does not directly measure how difficult that text is for a language model to process. To capture this second dimension, we analyze \emph{surprisal}, operationalized as perplexity under next-token prediction. Perplexity reflects the model’s uncertainty and sensitivity to lexical frequency, syntactic structure, and register, making it a natural measure of \emph{comprehension difficulty}.

If tokenization cost were a reliable proxy for comprehension, texts with similar token inflation should exhibit similar perplexity increases. The following analysis tests this assumption directly.

We compute sentence-level perplexity using the Qwen 2.5 1.5B language model.\footnote{We used the base checkpoint \texttt{Qwen/Qwen2.5-1.5B} via Hugging Face to ensure a raw probability distribution unaffected by instruction tuning or alignment bias.} This model was chosen for its strong multilingual performance and the availability of a matching tokenizer for direct cost comparisons. For each dataset, we compare perplexity on the historical text to that of a semantically equivalent modernized version, and report perplexity ratios (historical / modern). This paired design controls for content and length, allowing differences to be attributed to linguistic form rather than topic.

Unless stated otherwise, perplexity ratios are reported as the mean of per-sentence ratios (paired historical/modern sentences), which is more robust to heavy-tailed perplexity distributions than the ratio of aggregate means.

Table~\ref{tab:perplexity} reports perplexity values and historical-to-modern ratios across datasets under the no-context condition. These results reveal substantial variation in how surprising historical text is to modern language models, even when tokenization costs are comparable. Manzoni (Quarantana) shows minimal surprisal. The 19th-century Italian control condition exhibits only a modest increase in perplexity relative to contemporary Italian (1.70×). This confirms that canonical, standardized Italian—despite being historically marked—behaves as a best-case historical baseline rather than a genuine stress test.

Russian shows moderate surprisal despite severe tokenization. Russian 18th-century texts are only marginally more surprising than their modern equivalents (1.25×), even though tokenization inflation reaches up to 36\%. This indicates that orthographic disruption alone does not necessarily translate into high predictive uncertainty.

The 17th-century Italian corpus exhibits a mean perplexity ratio of \textbf{2.42$\times$}, with the academic \textit{Miscellanea} text reaching 3.22$\times$. This represents the largest comprehension gap observed in the study and reflects substantial lexical and syntactic distance from modern Italian.

Taken together, these results demonstrate that texts with similar tokenization penalties can differ significantly in surprisal.

\begin{table}[t]
\centering
\caption{Cross-dataset perplexity comparison under the no-context condition. Mean PPL values are reported for reference; ratios are the mean of per-sentence historical/modern ratios.}
\label{tab:perplexity}
\begin{tabular}{l l r r}
\hline
Dataset & Condition & PPL (mean) & Ratio \\
\hline
Italian 17th c. Corpus (1610--1689) & Historical & 115.5 & \textbf{2.42$\times$} \\
(n=732) & Modern & 51.8 & 1.00$\times$ \\
\hline
Russian civil print (18th c.) & Historical & 38.2 & 1.25$\times$ \\
(n=374) & Modern & 36.7 & 1.00$\times$ \\
\hline
Manzoni \textit{Quarantana} (1840) & Historical & 41.5 & 1.70$\times$ \\
(n$\approx$340) & Modern (LLM) & 26.6 & 1.00$\times$ \\
\hline
\end{tabular}
\end{table}

\begin{figure}[t]
  \centering
  \includegraphics[width=\linewidth]{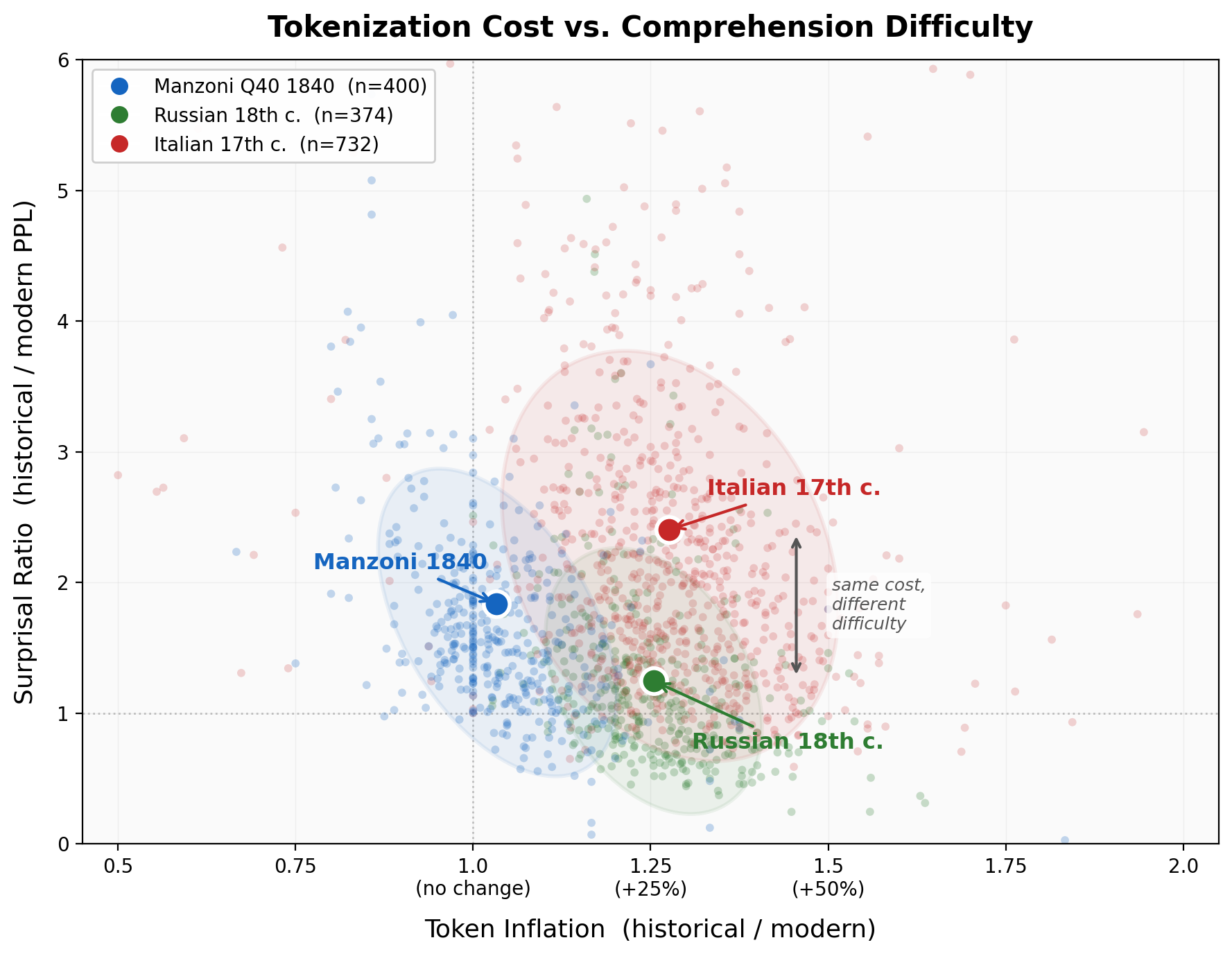}
  \caption{\textbf{Dissociation of Encoding Cost and Predictive Difficulty.} 
  Comparing token inflation (x-axis) vs. surprisal ratio (y-axis) reveals distinct historical regimes. 
  \textbf{Russian 18th c. (Green)} and \textbf{Italian 17th c. (Red)} share similar encoding costs ($\approx$+25--30\% inflation due to orthography), yet their predictive difficulty diverges sharply. 
  Russian imposes a ``tokenization tax'' without confusing the model (low surprisal), whereas 17th-century Italian incurs both high cost and high uncertainty, reflecting a deeper distributional mismatch. 
  \textbf{Manzoni (Blue)} serves as a high-exposure baseline with minimal drift.}
  \label{fig:cost_vs_surprisal}
\end{figure}

\textbf{Tokenization Cost vs. Comprehension Difficulty.} The contrast between Russian and 17th-century Italian is particularly instructive (see Figure~\ref{fig:cost_vs_surprisal}). 
Both datasets incur similar tokenization penalties ($\approx$25--30\%), yet their perplexity behavior diverges sharply. Russian shows only a modest surprisal increase, whereas the 17th-century Italian corpus is on average 2.4$\times$ more surprising than its modernized counterpart (rising to over 3.2$\times$ for the academic 1687 text).

This dissociation---clearly visible as the vertical gap in Figure~\ref{fig:cost_vs_surprisal}---indicates that \textit{tokenization inefficiency is neither necessary nor sufficient to explain comprehension difficulty}. Instead, perplexity is dominated by factors such as vocabulary rarity, syntactic structure, and register. In the 17th-century texts, archaic lexical items (e.g., \textit{oggidì}, \textit{eziandio}, \textit{Imperocchè}) and Latin-influenced syntax introduce strong distributional mismatches that persist even when orthographic normalization is applied.

Conversely, in Russian, much of the historical difficulty is concentrated at the orthographic level. Once characters are normalized, the underlying lexical and syntactic patterns remain relatively familiar to the model, limiting surprisal despite severe token fragmentation.

\textbf{Distributional Effects and Genre Sensitivity.} The distributional spread in Figure~\ref{fig:cost_vs_surprisal} further distinguishes surface noise from linguistic difficulty. 
While the Russian data clusters tightly, indicating uniform difficulty, the 17th-century Italian corpus exhibits high variance and a heavy right skew. 
Table~\ref{tab:genre_breakdown} reveals that this variance is driven by register rather than time. 
Token Inflation remains constant across the century ($\approx$+26--30\%), confirming that the ``orthographic tax'' (long s, V/U) is a stable baseline cost.

However, surprisal fluctuates independently of this cost. 
The highest contrast appears between the two newest texts: the academic \textit{Miscellanea} (1687) is the hardest in the corpus (3.22$\times$), while the nearly contemporary \textit{Ritratti} (1689) is the easiest (1.81$\times$). 
Correlation analysis supports this dissociation: within individual books, we observe a weak negative correlation between token inflation and perplexity ratio ($r$ ranges from $-0.13$ to $-0.43$), confirming that the sentences which are most expensive to encode are not necessarily the ones the model finds most difficult to understand. 

As shown in Table~\ref{tab:qualitative_examples}, this difficulty is often mediated by structural completeness. The model successfully predicts long, rhythmic theological passages despite frequent long-s usage (PPL 17.1), but faces extreme uncertainty (PPL > 1700) with short, context-poor fragments containing abbreviations or dense enclitics (\textit{Aggiungan{\longs}i}), typical of the academic register.

\begin{table}[t]
\centering
\caption{\textbf{The Genre Effect:} Breakdown of the 17th-century corpus (ordered by date). While token inflation (encoding cost) is nearly identical across the century ($\approx$+27--30\%), surprisal (predictive difficulty) varies dramatically by genre. Notably, the hardest text (1687) and easiest text (1689) are nearly contemporary, ruling out chronological drift as the primary driver.}
\label{tab:genre_breakdown}
\begin{tabular}{l l l r r}
\hline
Year & Genre & Text ID & Token Infl. & PPL Ratio \\
\hline
1610 & Religious Treatise & \textit{Simbolo} & +26.8\% & 2.44$\times$ \\
1673 & Biography & \textit{Achillini} & +30.3\% & 2.10$\times$ \\
1674 & Historical Narrative & \textit{Historica} & +26.4\% & 1.86$\times$ \\
1687 & Academic Preface & \textit{Miscellanea} & +28.2\% & \textbf{3.22$\times$} \\
1689 & Descriptive Portraits & \textit{Ritratti} & +28.9\% & \textbf{1.81$\times$} \\
\hline
\end{tabular}
\end{table}

\begin{table}[t]
\centering
\small
\caption{\textbf{The Context Factor:} Qualitative analysis reveals that sentence length and structural completeness are major drivers of surprisal. The model successfully predicts long, rhythmic theological prose (providing its own internal context) even with heavy orthographic noise. Conversely, it struggles with short, context-poor fragments typical of the academic register, especially those containing abbreviations or segmentation artifacts.}
\label{tab:qualitative_examples}
\begin{tabular}{p{0.70\linewidth} r r}
\hline
\textbf{Sentence Text} & \textbf{Length} & \textbf{PPL} \\
\hline
\multicolumn{3}{l}{\textit{Easiest (High Internal Context)}} \\
``Et perciò non \textbf{\longs}olamente con frutto maggiore, ma etiandio cõ maggior diletto mirano le per\textbf{\longs}one \textbf{\longs}pirituali que\textbf{\longs}te co\textbf{\longs}e create...'' (1610) & 239 chars & 17.1 \\
\multicolumn{3}{l}{\textit{[Long, redundant theological rhetoric makes prediction easy despite `etiandio', `cõ', `{\longs}']}} \\
\hline
\multicolumn{3}{l}{\textit{Hardest (Low Context / Fragmentation)}} \\
``Aggiungan\textbf{\longs}i a mentouati Mon\textbf{\longs}ig.'' (1687) & 31 chars & 2697.5 \\
\textit{[Abbreviation + Enclitic; lacks context]} & & \\
``Chi loderà degnamente que\textbf{\longs}to che dentro à noi altri viue?'' (1610) & 57 chars & 1705.5 \\
\textit{[Short rhetorical question; archaic `viue' + `que{\longs}to']} & & \\
\hline
\end{tabular}
\end{table}

A potential confound is that full LLM modernization may produce unusually ``easy'' modern Italian, inflating historical-to-modern perplexity ratios. To control for this, we evaluate a conservative normalization on the hardest subset (\textit{Miscellanea} 1687) that changes only orthography ({\longs}$\rightarrow$s) while preserving historical vocabulary and syntax.

Table~\ref{tab:modernization_control} shows that this conservative variant is \emph{more} surprising than the original historical text (3.63$\times$ vs. 3.27$\times$ relative to the fully modernized baseline). This indicates that surprisal is dominated by vocabulary and syntax rather than orthography alone, and that partial normalization can be counterproductive.

Why would removing historical orthography \emph{increase} perplexity? We hypothesize that this reflects a distributional mismatch: the original text contains consistent historical signals ({\longs}, archaic vocabulary, Latin-influenced syntax) that may collectively prime the model toward a historical register. Normalizing only {\longs}$\rightarrow$s creates a hybrid form that looks superficially modern but retains archaic vocabulary, violating the model's expectations for contemporary Italian without triggering whatever implicit accommodation it may have for consistently historical text.

\begin{table}[t]
\centering
\caption{Modernization conservativeness control for the \textit{Miscellanea} subset (1687). Ratios are computed relative to the fully modernized baseline.}
\label{tab:modernization_control}
\begin{tabular}{l r r}
\hline
Version & PPL (mean) & Ratio vs. modern \\
\hline
Original (1687) & 174.8 & 3.27$\times$ \\
Conservative ({\longs}$\rightarrow$s only) & 194.0 & \textbf{3.63$\times$} \\
Modern (LLM) & 53.4 & 1.00$\times$ \\
\hline
\end{tabular}
\end{table}

The perplexity analysis yields three central conclusions. 
First, tokenization inefficiency does not predict model confusion. Russian and 17th-century Italian share the same ``encoding tax'' ($\approx$+28\%), yet their surprisal levels diverge, indicating that difficulty is driven by register and syntax rather than spelling. 
Second, historical difficulty is not a linear function of time. The easiest and hardest texts in our 17th-century corpus were published only two years apart (1687 vs. 1689), proving that genre (academic vs. descriptive) outweighs temporal drift. 
Finally, canonical texts can be misleading. 19th-century literature (Manzoni) behaves as a high-exposure baseline, masking the structural challenges posed by non-canonical, pre-standardized text.

In the next section, we show that this difficulty is partially reducible: a minimal temporal context prompt can cut surprisal by nearly half, demonstrating that the observed gap often arises from domain mismatch rather than irreducible linguistic distance.

\section{Context Sensitivity and Mitigation}

The perplexity results in Section~6 suggest that historical surprisal arises not just from linguistic distance, but from a mismatch between the model's training distribution and the specific historical register.
If so, explicitly signaling the temporal domain should reduce surprisal by priming the model's expectations.

We tested a minimal, model-agnostic intervention: prepending a short context string to each historical sentence indicating its language and period (e.g., ``Il seguente è un testo italiano del XVII secolo''). The text itself remained unchanged.

For the 17th-century Italian corpus, the effect is significant: perplexity drops by an average of 58\% (ranging from $-$43\% to $-$66\% across texts).
For the narrative and theological texts (1610, 1673, 1674, 1689), the reduction is consistent ($\approx$65\%). 
The academic \textit{Miscellanea} (1687) shows a smaller but still substantial improvement ($-$43\%), consistent with our earlier finding that its difficulty stems partly from structural fragmentation which temporal priming alone cannot resolve.

In contrast, Russian 18th-century text shows a more modest reduction ($-$17\%), though this is sufficient to lower its perplexity below that of the modern baseline. This disparity suggests that context priming is most essential for texts with high distributional divergence (Italian 17th c.), whereas texts with primarily orthographic difficulty (Russian) benefit less because the model is already robust to character-level noise.

These results identify temporal context prompts as a high-value, cost-free mitigation strategy. By simply informing the model of the document's date, digital library systems can reduce predictive uncertainty by more than half for early modern texts, without the need for expensive normalization.

\section{Semantic Robustness Across Time}

High perplexity indicates predictive difficulty, but it does not necessarily imply semantic failure. 
A key question for digital libraries is whether this generative confusion compromises downstream tasks like search and retrieval.
To answer this, we assess whether historical and modern variants remain aligned at the representational level by measuring embedding similarity between aligned sentence pairs ($n=732$).

To ensure robustness, we validated these results across three distinct architectures: OpenAI (\texttt{text-embedding-3-large}), Google's multilingual LaBSE, and the hidden states of Qwen 2.5-1.5B (the same model used for perplexity analysis).

Table~\ref{tab:embedding} summarizes the results. Across all conditions, similarity remains high ($\geq$0.85), substantially exceeding random baselines (0.22--0.35). 
Qwen 2.5 achieves the highest semantic alignment (0.938) of all models tested.
So the very same model that faces high predictive uncertainty (Section 6) typically maintains a near-perfect semantic representation of the historical input. 
In effect, the model ``sees through'' the orthographic noise (e.g., long s) in its internal states, even if generating those tokens is probabilistically expensive.

\begin{table}[t]
\centering
\caption{Cross-dataset embedding similarity (Cosine Similarity). Qwen 2.5 (generative) demonstrates superior semantic stability compared to dedicated embedding models, despite high perplexity.}
\label{tab:embedding}
\begin{tabular}{l l r r r}
\hline
Dataset & Time gap & OpenAI & LaBSE & Qwen 2.5 \\
\hline
Russian (18th c.) & $\sim$250 yr & 0.909 & 0.945 & 0.976 \\
Manzoni \textit{Quarantana} & $\sim$180 yr & 0.915 & 0.926 & 0.987 \\
Italian 17th c. Corpus & 335--415 yr & 0.877 & 0.850 & 0.938 \\
\hline
\end{tabular}
\end{table}

To understand what drives embedding degradation, we examine correlations between similarity and other metrics across the Italian 17th-century dataset (Table~\ref{tab:correlation}). Token inflation shows effectively no correlation with semantic similarity ($r=+0.12$), whereas perplexity ratio shows a moderate negative correlation ($r=-0.29$).

\begin{table}[t]
\centering
\caption{Correlation between embedding similarity (OpenAI) and other metrics (Italian 17th c. Corpus, $n=732$).}
\label{tab:correlation}
\begin{tabular}{l r}
\hline
Metric pair & Pearson $r$ \\
\hline
Similarity vs. Token Inflation & $+0.12$ \\
Similarity vs. Perplexity Ratio & \textbf{$-0.29$} \\
Similarity vs. Sentence Length & $+0.22$ \\
\hline
\end{tabular}
\end{table}

These results reinforce our central finding: \emph{tokenization efficiency is a poor proxy for semantic robustness}. 
The fact that encoding cost (inflation) is uncorrelated with embedding quality confirms that models can successfully map archaic orthography to the correct semantic space, even when they struggle to predict it syntactically.

For digital library applications, this dissociation is encouraging. It suggests that while historical text may be expensive to encode (tokens) and hard to generate (perplexity), it remains highly accessible for retrieval and indexing tasks without the need for aggressive normalization.

\section{Discussion}

This study set out to disentangle what makes historical Italian “difficult” for modern language models. By jointly analyzing tokenization efficiency, perplexity, embedding similarity, and context sensitivity, we show that historical difficulty is not a unitary phenomenon, but a composite of distinct costs.

\paragraph{Tokenization is a mechanical tax, not a proxy for difficulty.}
Our results demonstrate that encoding cost and comprehension difficulty are effectively uncorrelated. Across the 17th-century corpus, token inflation remained constant ($\approx$28\%) regardless of genre or difficulty. This confirms that tokenization acts as a mechanical ``orthographic tax''—driven by the long s—rather than a signal of linguistic complexity. Consequently, metrics based on subword fertility or encoding length are misleading proxies for how well a model understands historical text.

\paragraph{Genre outweighs chronology.}
Historical difficulty is not a linear function of time. The most notable dissociation in our dataset occurred between the academic \textit{Miscellanea} (1687) and the descriptive \textit{Ritratti} (1689). Despite being published only two years apart, the academic text was nearly twice as surprising to the model. This demonstrates that \emph{register}—specifically the structural fragmentation of academic prose versus the rhythmic redundancy of narrative—is a stronger determinant of model performance than chronological age.

\paragraph{Comprehension persists despite uncertainty.}
While the Qwen model exhibited high predictive uncertainty (perplexity > 100), its internal embeddings of the same text maintained near-perfect semantic alignment with modern Italian (Cosine Sim = 0.938). This suggests that LLMs can successfully project archaic surface forms into the correct semantic space even when they lack the probabilistic confidence to generate them. For digital libraries, this implies that search and retrieval pipelines may be far more robust to historical noise than generative tasks.

\paragraph{Canonical texts are misleading benchmarks.}
Manzoni’s \textit{Quarantana} behaved as a best-case baseline rather than a stress test, exhibiting minimal surprisal and perfect semantic alignment. This confirms that models trained on web-scale data likely have verbatim or near-verbatim exposure to canonical literature. Evaluations that rely exclusively on such high-resource texts risk systematically underestimating the challenges posed by the ``long tail'' of non-canonical, digitised heritage materials.

\paragraph{Context prompting as a diagnostic tool.}
The effectiveness of minimal temporal context prompts (reducing surprisal by $\approx$60\%) highlights that much of the observed ``difficulty'' is simply domain mismatch. When the model is primed to expect 17th-century Italian, it adapts its probability distribution significantly. However, the residual gap in the academic 1687 text indicates that prompting cannot fix structural issues: while context helps the model anticipate archaic words, it cannot compensate for the lack of redundancy in fragmented, abbreviated lists.

Taken together, these findings advocate for a decoupled evaluation framework. Effective historical NLP must distinguish between \emph{encoding cost} (solved by vocabulary expansion), \emph{predictive uncertainty} (mitigated by prompting), and \emph{semantic robustness} (already high), rather than treating ``historical language'' as a single, insoluble problem.

\section{Conclusion}

This study decomposed the difficulty of historical Italian for modern language models, revealing that ``historical difficulty'' is not a monolithic barrier but a composite of interacting factors: encoding cost, predictive uncertainty, and semantic robustness. Our results demonstrate that tokenization inefficiency, while imposing a consistent computational tax ($\approx$28\% overhead), is neither a necessary nor sufficient predictor of model performance. The substantially larger surprisal gap observed in pre-1700 texts is driven primarily by register and syntactic structure rather than orthographic noise alone.

For digital library practitioners, these findings suggest a strategic division of labor when deploying LLMs on heritage collections.
We observe a notable dissociation where models struggle to predict historical text token-by-token (high perplexity) yet successfully project it into a coherent semantic space (high embedding similarity). 
Consequently, retrieval and discovery tasks---including semantic search, clustering, and classification---appear highly robust to historical variation and can likely be deployed without aggressive normalization. 
By contrast, generative tasks---such as OCR correction, summarization, or text generation---remain sensitive to high surprisal and require specific mitigation strategies.

To bridge this generative gap, we showed that minimal temporal context prompts (e.g., signaling the text's date) can reduce predictive error by over 60\%, offering a cost-effective intervention for library workflows. However, the persistence of residual surprisal in non-canonical texts underscores the risk of relying on high-exposure literary benchmarks like Manzoni’s \textit{I Promessi Sposi}. 
To accurately assess model suitability, libraries should evaluate performance not on standardized canons, but on the ``long tail'' of archival material where the genuine challenges of historical register reside.

Ultimately, effective historical text processing requires moving beyond surface-level metrics like token counts. By distinguishing between the mechanical cost of encoding and the semantic stability of representation, digital libraries can deploy language models where they are most reliable: not necessarily as perfect speakers of the past, but as highly capable readers of it.


\section*{Declaration on Generative AI}

During the preparation of this work, the author(s) employed Generative AI tools/services under the CEUR-WS activity taxonomy for the following purposes:

\begin{itemize}
 \item \textbf{Data generation / transformation:} 
  GPT-5-mini was used to generate modernized Italian 
  counterparts of 17th-century historical texts for 
  controlled comparative experiments. Prompts instructed 
  the model to preserve semantic content while updating 
  orthography, vocabulary, and syntax. All outputs were 
  manually verified for semantic fidelity.
  Additionally, Gemini 2.5 Pro and GPT-5-mini were used 
  for OCR transcription of 17th-century page images, 
  with outputs reviewed against the original scans.
  \item \textbf{Data analysis / representation:} 
  Pretrained language models (Qwen~2.5-1.5B, Gemma~3-1B, 
  Llama~3.2-3B, Mistral~Small-24B) were used to compute 
  perplexity. Embedding APIs (OpenAI, LaBSE, Qwen~2.5 
  hidden states) were used for semantic similarity analysis. 
  These served as numerical measurements only and did not 
  contribute to text generation or authorship.
 
\end{itemize}

No Generative AI tools were used to fabricate experimental results or to generate scientific claims. All reported analyses, metrics, and interpretations were produced and validated by the author(s), who take full responsibility for the integrity of the work.

\bibliography{bibfile}

\end{document}